\documentclass[letterpaper, 10 pt, conference]{ieeeconf}  
\IEEEoverridecommandlockouts   
\usepackage{cite}
\usepackage{algorithm} 
\usepackage{algorithmic}  
\usepackage[algo2e]{algorithm2e} 
\usepackage{amsmath,amssymb,amsfonts}
\usepackage{algorithm}
\usepackage{graphicx}
\usepackage{tcolorbox}

\usepackage{textcomp}
\def\BibTeX{{\rm B\kern-.05em{\sc i\kern-.025em b}\kern-.08em
    T\kern-.1667em\lower.7ex\hbox{E}\kern-.125emX}}
    
    \usepackage{lineno}
\usepackage{lettrine}

\usepackage{url}
\usepackage{moreverb}

\usepackage{graphicx}
\usepackage{subcaption}
\usepackage{float}
\usepackage{amsmath}
\usepackage{array}
\usepackage{multirow}
\graphicspath{ {Figures/} }

\usepackage{scalerel}
\usepackage{tikz}
\usetikzlibrary{svg.path}

\definecolor{orcidlogocol}{HTML}{A6CE39}
\tikzset{
  orcidlogo/.pic={
    \fill[orcidlogocol] svg{M256,128c0,70.7-57.3,128-128,128C57.3,256,0,198.7,0,128C0,57.3,57.3,0,128,0C198.7,0,256,57.3,256,128z};
    \fill[white] svg{M86.3,186.2H70.9V79.1h15.4v48.4V186.2z}
                 svg{M108.9,79.1h41.6c39.6,0,57,28.3,57,53.6c0,27.5-21.5,53.6-56.8,53.6h-41.8V79.1z M124.3,172.4h24.5c34.9,0,42.9-26.5,42.9-39.7c0-21.5-13.7-39.7-43.7-39.7h-23.7V172.4z}
                 svg{M88.7,56.8c0,5.5-4.5,10.1-10.1,10.1c-5.6,0-10.1-4.6-10.1-10.1c0-5.6,4.5-10.1,10.1-10.1C84.2,46.7,88.7,51.3,88.7,56.8z};
  }
}

\newcommand\orcidicon[1]{\href{https://orcid.org/#1}{\mbox{\scalerel*{
\begin{tikzpicture}[yscale=-1,transform shape]
\pic{orcidlogo};
\end{tikzpicture}
}{|}}}}

\usepackage{hyperref} 
\usepackage{dblfloatfix}    
\usepackage{svg}

\begin{document}
\title{\LARGE \bf
\textit{From Prompts to Pavement:} LMMs-based Agentic Behavior-Tree Generation Framework for Autonomous Vehicles}

\author{Omar~Y.~Goba$^{1,2}$, Ahmed~Y.~Gado$^{1,2}$, \\ Catherine~M.~Elias$^{1,2\orcidicon{0000-0002-1444-9816}\,}$,~\IEEEmembership{Member,~IEEE} and Ahmed Hussein$^{3}$,~\IEEEmembership{Senior Member,~IEEE}%
\thanks{*This work was not supported by any organization}%
\thanks{$^{1}$Computer Science \& Engineering Department, German University in Cairo (GUC), Egypt {\tt\small omaryasserassem1@gmail.com, ahmed.yahia.gado@gmail.com, catherine.elias@ieee.org} }%
\thanks{$^{2}$C-DRiVeS Lab: Cognitive Driving Research in Vehicular Systems, Cairo, Egypt {\tt\small cdrives.researchlab@gmail.com}}%
\thanks{$^{3}$IAV GmbH, Berlin, Germany {\tt\small ahmed.hussein@ieee.org}}%
}

\markboth{Journal of \LaTeX\ Class Files,~Vol.~14, No.~8, August~2015}%
{author1 \MakeLowercase{\textit{et al.}}:title here}
%



\maketitle
\begin{abstract}
Autonomous vehicles (AVs) require adaptive behavior planners to navigate unpredictable, real-world environments safely. Traditional behavior trees (BTs) offer structured decision logic but are inherently static and demand labor-intensive manual tuning, limiting their applicability at SAE Level 5 autonomy. This paper presents an agentic framework that leverages large language models (LLMs) and multi-modal vision models (LVMs) to generate and adapt BTs on the fly. A specialized Descriptor agent applies chain-of-symbols prompting to assess scene criticality, a Planner agent constructs high-level sub-goals via in-context learning, and a Generator agent synthesizes executable BT sub-trees in XML format. Integrated into a CARLA+Nav2 simulation, our system triggers only upon baseline BT failure, demonstrating successful navigation around unexpected obstacles (e.g., street blockage) with no human intervention. Compared to a static BT baseline, this approach is a proof-of-concept that extends to diverse driving scenarios.
\end{abstract}


%
\section{Introduction}\label{sec1}
\PARstart{T}{he} field of autonomous vehicles (AVs) has been one of the fastest growing areas of research. Organizations such as SAE International have established an international categorization of automation, ranging from Level 0 to Level 5\cite{sae2021taxonomy}. Recent attempts have been made to bridge the gap from level 4 (L4) to level 5 (L5), yet progress remains limited. SAE L4 describes a vehicle that can drive itself in both a limited and known environment, conversely L5 vehicles can drive themselves in both known and unknown environments. This distinction implies that the latter has the ability for adaptable behavior planning.

Behavior planning in the context of AVs is defined as the sequence of high level actions that the vehicle should take in response to its environment and goal. Behavior planning has traditionally been defined on statically defined behavior trees (BTs), rule-based systems, Finite State Machines, or learning based approaches like Reinforcement Learning. BTs in particular have been the focal point in the industry, as they are a powerful tool for behavior planning in AVs\cite{conejo2024behavior,kang2025decision,kang2022behavior}.

BTs are a mathematical model for encoding decision logic. Their hierarchical structure allows them to represent complex tasks as compositions of simpler sub-tasks. They are composed of nodes which can be either \textit{composite} or \textit{leaf}. Composite nodes are responsible for controlling the flow of execution; Some prime examples, but not limited to, are \textit{Selector}, \textit{Sequence}, and \textit{Parallel}. Leaf nodes, on the other hand, are the actual actions, or conditions, that the vehicle can take\cite{chen2024Integrating, styrud2022combining, colledanchise2019towards}. However, these advantages come at a cost; BTs are inherently static.

Despite the maturity of BT based behavior planning, they remain limited by their static nature. The rigidity of BTs is a significant limitation, as they require to be manually tuned by experts based on the specific environment and task. This manual tuning process has two main caveats; Firstly it is time consuming and labor intensive. This is due to the fact that the tuning process requires a deep understanding of the environment and task. The second caveat is that the tuning process is only effective in known environments, and has the potential to fail in an unknown environment. Such limitations make BTs ill-suited for deployment in L5 context, as L5 requires a vehicle to be able to adapt to a dynamic environment.

Achieving true L5 has a compounding effect on a myriad of real world facets. An adaptable behavior planner significantly aids safety, by providing a robust handling of unpredictable scenarios. Furthermore, L5 is a technological leap that enables scalability and practical application across diverse driving conditions.

\section{State-of-the Art}\label{sec2}
To leverage said rigidity, we explore \textit{Large Language Models} (LLMs), which have emerged as a powerful tool for \textit{natural language processing} (NLP) tasks. They have proven to be a capable tool for generating text, answering questions, and even performing reasoning tasks\cite{vaswani2017attention, wei2022chain}. LLMs are a compelling candidate for the task of generating BTs. This arises from the fact that BTs are a form of structured text, which makes the idea of generating a BT a form of a NLP task. LLMs have an outstanding ability for adaptability by leveraging \textit{in Context Learning} (ICL) to generate text based on a given prompt. The NLP community has seen a surge in developing novel techniques to leverage ICL\cite{brown2020language, kojima2022large}. The benefit of using ICL is that it allows the model to adapt to a new task without the need for retraining. This further supports the argument that LLMs are a natural fit for the task of generating BTs, as they can be prompted to generate BTs based on a given context, in this situation the context will be the environment and a task, without being retrained.

Following the immense success of LLMs in text based tasks, the field of computer vision adapted LLMs to vision based tasks. This led to the creation of Large Multi-modal Models (LMMs). LMMs are capable of processing both text and image data, allowing them to perform tasks that require both modalities. This is particularly relevant for AVs, as they rely heavily on visual data from cameras and sensors to understand their environment\cite{dosovitskiy2020image}.

As of 2025, the field of LLMs have seen a dramatic transformation from monolithic architectures to more flexible, agentic architectures. This shift stems from the growing need to handle increasingly complex tasks. Agentic architectures are based on multi-agent systems (MAS), which enable the definition and coordination of specialized agents, each responsible for a specific task\cite{ao2024llm, wang2024describe, xu2024magic}. Such complexity necessitates effective task decomposition. MAS offers a structured way to achieve this and, crucially, allows for integrating behavior trees (BTs) with a greater degree of adaptability for more complex tasks.

As mentioned previously, BTs are a powerful tool for behavior planning in AVs, and have been used extensively in the industry\cite{chen2024Integrating, styrud2022combining, colledanchise2019towards}. However, BTs require manual tuning and lack the necessary adaptability to handle the dynamic nature of real-world environments. BTs have been attempted to imbue with a degree of adaptability by using genetic algorithms, reinforcement learning, or knowledge graphs\cite{conejo2024behavior, kang2022behavior, kang2025decision}. However, these approaches still suffer from one or more of the following: a reliance on a predefined structure, a large number of manually defined subtrees, or a large amount of training data. Meanwhile, a novel approach was introduced to use LLMs and LVMs to generate BTs. Researchers have found promising results in the field of robotics, but this has not yet been explored in the field of AVs\cite{ao2024llm, cai2024hbtp, lykov2023llm, cao2023robot, zhou2024llm}. Finally, there have been great leaps in the diverse applications of MAS. Although once again, these have not yet been explored in the field of AVs or generated BTs\cite{ao2024llm, wang2024describe, xu2024magic}. Given all that, it becomes clear that there is a multi-domain convergence of using MAS, leveraging LLMs and LVMs, to generate BTs for behavior planning in AVs. This presents a compelling gap in the literature.

Therefore, this paper introduces \textit{From Prompts to Pavement}, a framework that leverages an ensemble of both LLMs and LVMs to generate BTs for AVs.
\begin{enumerate}
	\item {
	      A failure-triggered mechanism was developed to coordinate agent activation.
	      }\item{
	      A vision-based criticality assessment was created using LVMs and spatial reasoning.
	      }\item{
	      A plan synthesizer and BT generator were constructed using ICL and LLMs to produce adaptable BTs.
	      }\item{
	      A high-fidelity simulation environment was assembled through the integration of CARLA with Nav2, enabling real time behavior evaluation.
	      }
\end{enumerate}

\begin{figure*}[t]
    \centering
    \includegraphics[width = 1\linewidth]{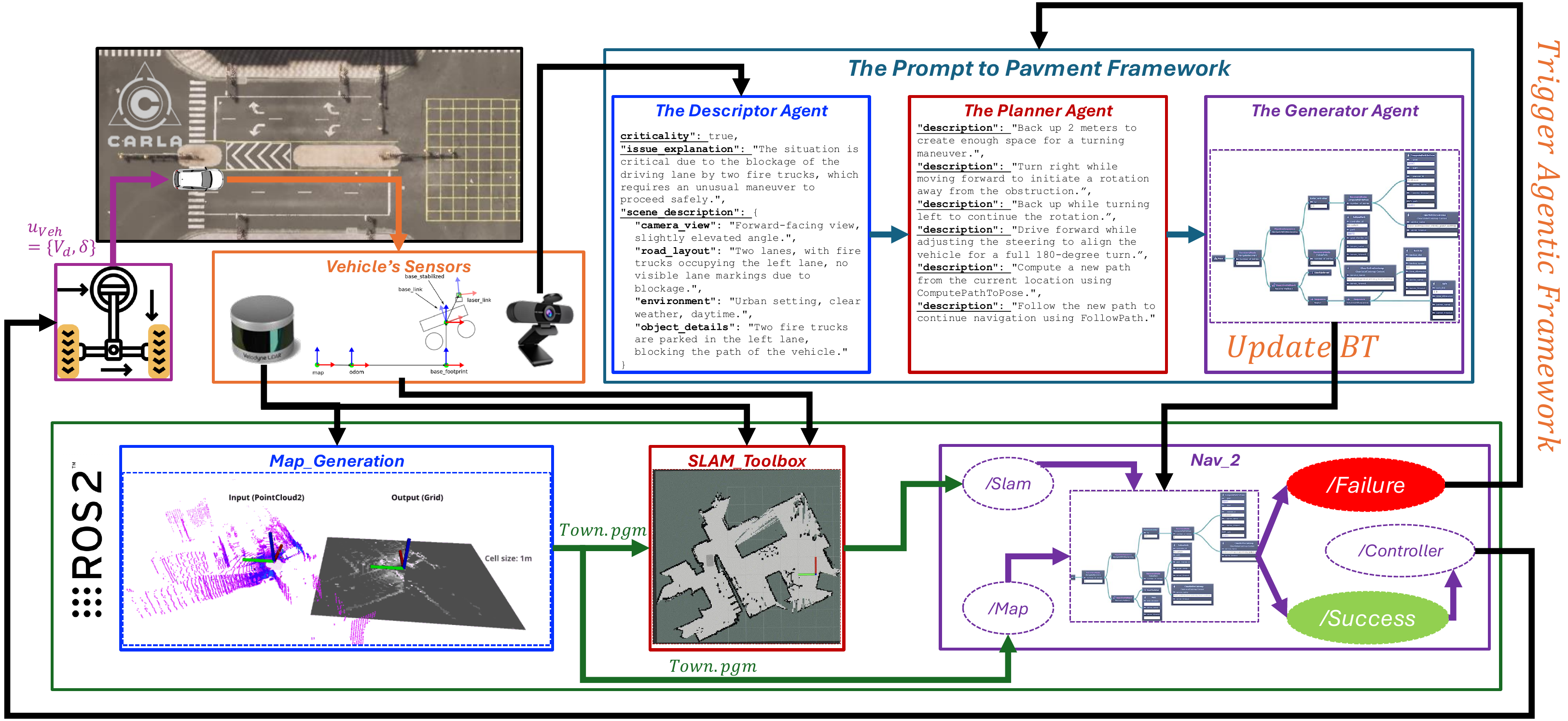}
    \caption{The Overall ROS-Based Autonomous Driving Stack with the Agentic Behavior-Tree Generation Framework}
    \label{fig:SysArch}
\end{figure*}

\section{Methodology}\label{sec3}
\subsection{System Overview}\label{sec3.a}
In the proposed system, a ROS-based architecture has been designed to mimic the full autonomous driving stack. This stack uses a \texttt{ROS2 Humble} \cite{doi:10.1126/scirobotics.abm6074} with \texttt{publisher-subscriber} communication protocol among the different included \texttt{ROS Nodes} to facilitate the exchange of necessary \texttt{ROS Messages}. 

The designed driving stack is mainly composed of three different layers: \textbf{\textit{SLAM Layer}}, \textbf{\textit{Perception Layer}}, and \textbf{\textit{Navigation and Control Layer}}. In order to empirically test the aforementioned architecture's effectiveness, a high-fidelity simulation environment was constructed atop \texttt{CARLA}\cite{Dosovitskiy17}

\subsubsection{Simulation Environment Preparation}
Inorder to facilitate the use of \texttt{CARLA} with \texttt{ROS2}, \texttt{carla\_ros\_bridge} package has been used. During the testing, an ego vehicle was spawned, equipped with the following sensors that enables the driving stack layers:
\begin{itemize}
    \item Odometry data was used to estimate the vehicle’s position and velocity.
    \item A LiDAR with a 50m range, 64 channels, FOV (-26.8 to 2.0 degrees).
    \item An RGB front camera with 1920x1080 resolution, and a FOV of 90.0 degrees.
\end{itemize}
The specifications of these sensors are set to match the commercial sensors inserted in the currently commercially available L3-L4 autonomous vehicles. This will help mimicking reality inside the simulation environment.

\subsubsection{SLAM Layer}
First, the \texttt{CARLA} map was generated using Simultaneous Localization and Mapping (SLAM) algorithm. This was done using the \texttt{slam\_toolbox} package \cite{Macenski2021} and was populated by manually driving the car through the environment. As the car was driven, the LiDAR point cloud readings were converted to laser scans and published to the \texttt{SLAM} node.

\subsubsection{Perception, Navigation and Control Layers}
\texttt{Nav2}\cite{macenski2020marathon2} pipeline is utilized in this architecture. This is mainly due to the fact that it uses behavior-trees in the planning. This is very pivotal since it can be considered as the baseline upon which the agentic BT generation can rely on. 

The \texttt{Nav2} was configured with \texttt{CARLA} by connecting the necessary topics via \texttt{ROS 2}. This was followed by manually setting and validating the \texttt{Nav2} parameters. Some examples of the parameters are the max/min velocities, and localization parameters. All parameters are configured to respect the configuration of the Ackermann steering, vehicle dimensions constraints. 

The \texttt{Nav2} package basically subscribes the \texttt{SLAM} node to have the prior information about the generated global map. Also, it inherits the perception layer where the LiDAR is used mainly for obstacle detection which is essential for the navigation to handle the obstacle collision scenarios. It requires taking a goal pose from the user to drive the vehicle to. 

Afterward, the BT is traversed, activating the corresponding nodes matching the instant scenario, till reaching the final action nodes. These final action nodes are considered the main trigger to the proposed agentic behavior-tree generation framework in case of failure.

In case of a successful action by the tree, the \texttt{Nav2} starts the \texttt{Navigate To Pose}, generating sequence of waypoints to be followed using the selected controller. As for the control of the vehicle, the Model Predictive Path Integral (MPPI) Controller is generate the proper control action to drive the vehicle including throttle and steering.

\subsubsection{Overall System Flow}
To better study this agentic pipeline introduced in Section \ref{sec3.b}, the triggering condition must be explicitly defined. The vehicle will have an onboard baseline behavior tree defined by \texttt{Nav2}. This tree will utilize the path planner to aid in controlling the behavior of the car. 

In the condition that the root node of this baseline tree returns \texttt{Failure}, the entire agentic pipeline initiates execution. The scenes that can result in to a \texttt{Failure} are ones that the engineer who developed said baseline tree could not anticipate. As an example a prolonged lane abstraction is categorized as an unpredictable scene. 

This further motivates the idea of adaptability. To put more formally;
\begin{equation}
    \mathcal{B}_{base}: \mathcal{E} \to \{\texttt{Success}, \texttt{Failure}\}
\end{equation}

Where $\mathcal{B}_{base}$ is the baseline behavior tree, $\mathcal{E}$ is the set of all possible scenes, and $\{\texttt{Success}, \texttt{Failure}\}$ is the set of possible statuses. 

The triggering condition is then defined as:
\begin{equation}
    \mathcal{B}_{base}(e) = \texttt{Failure} 
\end{equation}
Fig. \ref{fig:SysArch} illustrates the overall ROS-Based autonomous driving stack adopting the proposed agentic Behavior-Tree generation framework known as \textbf{\textit{"Prompt to Pavement"}}.

\subsection{Agentic Behavior-Tree Generation Framework} \label{sec3.b}
To ensure the adaptability of BTs, a 3-agent MAS architecture is developed. These agents are: the \textit{\textbf{Descriptor}}, the \textit{\textbf{Planner}}, and the \textit{\textbf{Generator}}. Each agent is chained to the other in a sequential manner as mathematically represented in \eqref{eq:m}, where the output of one agent is fed as input to the next agent. 
\begin{gather} 
\textbf{Descriptor:} \quad \mathcal{I} \to \mathcal{O} \label{eq:m_a} \nonumber
\\
\textbf{Planner:} \quad \mathcal{O} \to \mathcal{P}  \label{eq:m_b} \nonumber
\\
\textbf{Generator:} \quad \mathcal{G} \to \mathcal{B} \label{eq:m_c} \nonumber \\
\textbf{pipeline} \triangleq \textbf{Generator} \circ \textbf{Planner} \circ \textbf{Descriptor} \label{eq:m}
\end{gather}
Where $\mathcal{I}$ is the set of all possible images,  $\mathcal{O}$ is the output of the \textbf{\textit{Descriptor}}, $\mathcal{P}$ is the high level plan, and $\mathcal{G} \in \mathcal{P}$ is the set of all possible sub-goals.

The \textit{\textbf{Descriptor}} is responsible for describing the environment and assessing the criticality of the situation. The second agent; the \textit{\textbf{Planner}}, takes the scene description provided by the first agent and constructs a high-level strategy on how the vehicle should behave. Finally, the \textit{\textbf{Generator}} takes said strategy and generates a BT.

A more rigorous definition of each agent will be found in the following sections. The pipeline aforementioned is an overview of the entire agent pipeline.

\subsubsection{The Descriptor Agent}
This agent serves as the initial stage of the pipeline. It uses an LMM as its model of choice. It takes as input an image $i$ captured from the vehicle's onboard RGB camera. It then initiates a Chain of Thought (CoT) to boost the agent's spatial reasoning accuracy. 

The flavor of CoT used was Chain of Symbols (CoS), where objects that are detected are given arbitrary symbols, and the physical relations constructed between each object are also given a symbol. 

This streamlines the token footprint and boosts the accuracy of the model in spatial related tasks\cite{hu2023chain}. Upon the completion of the reasoning process, the agent outputs a structured output in the format of \texttt{JSON} object. The primary keys in said object are: \texttt{isCritical}, \texttt{issueExplanation}, and \texttt{sceneDescription}.
\begin{gather}
    \textbf{Descriptor}\left(
i \in \mathcal{I} \equiv \mathbb{R}^{h \times w \times 3} 
\right) = o \\
    s.t.~~o  = \left\langle
\kappa \in \{0,1\},\;
e \in \Sigma^{*},\;
d \in \Sigma^{*}
\right\rangle
\in \mathcal O \nonumber
\end{gather}
Where $\kappa$ is the criticality key, $e$ is the issue explanation, and $d$ is the scene description. The symbols $\Sigma^{*}$ represent a string of arbitrary length. $\mathcal O$ is the set of all possible Observations produced by the \textit{\textbf{Descriptor}}, and $o$ is the \textbf{\textit{Descriptor}} output.

The first key, \texttt{isCritical}, is just a boolean value. In essence, this key acts as a filtering method. If the value is \texttt{False}, the entire pipeline halts. If the value is \texttt{True}, the pipeline continues to the next agent. This key computed by the \textbf{Descriptor} after the agent concludes the CoS stage. The model leverages the deep spatial analysis to determine if the scene is critical or not. 

To ground the model's output, \textit{Few Shot} is instrumental. A set of carefully selected examples is provided to the model, showcasing various critical and non-critical scenarios. This approach enhances the model's ability to accurately assess the criticality of the scene. \texttt{issueExplanation} key holds a natural language description of what is the actual issue shown in the image. Finally, the \texttt{sceneDescription}, holds a general description of the image. The final key is critical as it aids in grounding the next agents into the real situation and to prevent subsequent agents from hallucinating non-existent elements.

To ensure consistent and rigorous reasoning across inputs, the Descriptor agent was primed with a system prompt specifically engineered to elicit multi-step spatial analysis, validation, and hazard detection. The full prompt, expressed in markdown format, is provided in the illustrated prompt.
\begin{figure}
    \centering
    \begin{minipage}{1\linewidth}       
		\begin{tcolorbox}[colback=white!5!white,colframe=black!75!black,
				title=Descriptor Prompt]
			\vspace{1em}

			\textbf{Role}\\
			You are an image analysis agent analyzing images from a vehicle-mounted RGB camera to identify hazards and critical driving situations.

			\vspace{0.5em}
			\hrule
			\vspace{0.5em}

			\textbf{Tasks}\\
			Provide structured JSON output with four sections:

			\vspace{0.5em}
			\hrule
			\vspace{0.5em}

			\textbf{1. Chain-of-Thought (CoT) and Spatial Reasoning:}
			\begin{enumerate}
				\item Identify and list visible objects with positions and approximate distances.
				\item Classify pedestrians and cyclists as "vulnerable road users," noting activity and movement.
				\item Determine spatial relationships of vulnerable road users relative to lane markings and driving path.
				\item Identify potential hazards (users within 6 feet of lane or intersecting vehicle path).
			\end{enumerate}

			\vspace{0.5em}
			\hrule
			\vspace{0.5em}

			\textbf{2. Verification Step:}\\
			Perform self-check validations:
			\begin{itemize}
				\item Pedestrian/cyclist detection
				\item Spatial accuracy
				\item Hazard assessment
				\item Critical situation consistency
			\end{itemize}

			\vspace{0.5em}
			\hrule
			\vspace{0.5em}

			\textbf{3. Critical Situation Detection Criteria:}\\
			Situation is critical if:
			\begin{enumerate}
				\item Pedestrian/cyclist is on the road surface.
				\item Pedestrian/cyclist within 6 feet of driving lane.
				\item Trajectory intersects vehicle path.
				\item Indefinite or prolonged blockage (large vehicles, stalled cars, unsafe maneuvers required).
			\end{enumerate}
			Provide:
			\begin{itemize}
				\item Confidence level (0–100\%)
				\item "isCritical": true/false
				\item Step-by-step justification referencing CoT analysis.
			\end{itemize}

			\vspace{0.5em}
			\hrule
			\vspace{0.5em}

			\textbf{4. Description for Engineering Analysis:}\\
			Structured description including:
			\begin{itemize}
				\item Camera perspective
				\item Road conditions/layout
				\item Environmental context
				\item Precise locations, activities, trajectories of vulnerable road users
				\item Other relevant driving decision factors.
			\end{itemize}

			\vspace{0.5em}
			\hrule
			\vspace{0.5em}

			\textbf{Additional Guidelines:}
			\begin{itemize}
				\item Prioritize safety and caution.
				\item Clearly justify decisions referencing explicit criteria.
			\end{itemize}
			\vspace{1em}
		\end{tcolorbox}
	\end{minipage}
    \label{DescPrompt}
\end{figure}    
\subsubsection{The Planner Agent}
Once a detailed description of both the issue and the scene has been constructed by the \textbf{Descriptor}, the planner then starts developing a strategy. This agent uses a reasoning model as its LLM. 

The output of this agent is a structured sequence of sub-goals. Each sub-goal is written in plain English and is a high level step in resolving the given issue given the scene context. The choice of plain English as the encoding of the sub-goals was too boost interpretability and traceability. The sub-goals are assumed to be sequential and logically dependent. In other words, for the vehicle to follow said strategy, it must complete each step in order and all steps must result in a \texttt{success} status. 

This construction ensure alignment with the final output of this pipeline which will be a BT with a root node of type sequence.
\begin{gather}
    \textbf{Planner}(\mathcal{O})
= \mathcal{P} \equiv [\,g_1, g_2, \dots, g_n\,],
\quad
g_k \in \mathcal{G} \equiv \Sigma^{*} \\
\text{with } g_k \xrightarrow{\text{order}} g_{k+1}~:~\forall k,\;
\text{and }\bigwedge_{k=1}^n \mathrm{success}(g_k) \nonumber
\end{gather}
Where  $P$ is the high level plan, and $\mathcal{G}$ is the set of all possible sub-goals. The sub-goals are assumed to be ordered, where $g_k$ is the $k^{th}$ sub-goal.

\subsubsection{The Generator Agent}
Given the high level strategy provided by the \textbf{Planner}, this agent iterates over each sub-goal and constructs a sub-tree in \texttt{XML} format. The choice of using \texttt{XML} was necessitated on two fronts. Firstly using and Resource Description Framework (RDF) like XML, behavior tree representation becomes trivial. Secondly, \texttt{BehaviourTree.cpp} nattily supports \texttt{XML}. 

The \textbf{Generator} leverages a reasoning-enabled LLM to improve the alignment between the generated sub-trees and their respective sub-goals. Furthermore, this agent has been explicitly instructed via prompt engineering to utilize only the predefined leaf nodes made available by the path planner interface. This was achieved by using an array of carefully selected few-shots.
\begin{gather}
    \textbf{Generator}(g_i) = b_i \in \mathcal{B} \\
\mathcal{B}_{out} = \left\langle b_1, b_2, \dots, b_n \right\rangle \nonumber
\end{gather}
Where $g_i$ is the $i^{th}$ sub-goal, $\mathcal{B}$ is the set of all possible behavior trees, and $\mathcal{B}_{out}$ is the final output of the \textbf{Generator}. 

The final output of this agent is a sequence of behavior trees, where each tree corresponds to a sub-goal. This sequence will be be concatenated under a sequence node to produce the final behavior tree. Before execution, each sub-tree is manually reviewed to ensure it is well-formed XML, Any malformed sub-trees are identified and repaired by a human expert. 

\subsubsection{Agents' Configuration}
Regarding the agents' configuration, all three agents were configured to use a closed-source LMM provided by OpenAI. The model used was the \texttt{gpt-4o-mini}. The temperature was set for all agents to \textit{0.0} to ensure deterministic behavior during inference. 

Each  agent was instructed to return a structured output in the form of a \texttt{JSON} object to enable seamless handoff of structured information between agents. The prompt templates were customized per agent role but followed a shared structure to maintain coherence across the pipeline. 
\section{Results and Discussion}\label{sec4}
To evaluate the performance of the \textit{Prompt to Pavement} framework, it is assessed along three key dimensions: the accuracy of its scene interpretation (Descriptor), the generation efficiency of its agentic pipeline, and the behavioral quality of the generated BTs in simulation.

\subsection{Evaluation of scene interpretation (Descriptor)}\label{sec4a}
The core functionality of the descriptor agent is to assess the criticality of the scene. To evaluate this functionality, the dataset used was the \textit{Berkeley DeepDrive} (BDD-X) dataset\cite{kim2018textual}, for its $6000$ short videos of driving scenarios, each lasting $2 seconds$. 

Paired with each video is a human-generated description of both the action taken and the justification explaining the motivation behind the maneuver. The labels were used to generate a synthetic but rigorous approximation of the criticality of the scene.

To generate the criticality labels, a subset of 50 videos was selected from the dataset and annotated by a human expert to gauge the criticality of the scene. This subset became the ground truth for the criticality labels. 

An ensemble of three techniques was used to generate the criticality labels: \textit{Stacking Regressor}, \textit{Dense IR + weighted cosine}, and \textit{regex-based heuristic}. The final criticality label was generated by taking the weighted average of the three techniques. The weights—0.3, 0.4, and 0.3 respectively—were selected using grid search over a held-out test dataset to optimize performance. The choice for the ensemble was to ensure that the synthetic data module does not underfit due to the low number of labels.

The \textbf{\textit{Descriptor Agent}} was then tested on the 6000 videos (1 screenshot per video). The agent achieved a mean absolute error (L1 loss) of $0.5$ on a scale of $0$ to $1$ on the criticality labels. The L1 loss may not fully capture the binary decision performance of the descriptor agent. This relatively high loss indicates that an LLM based criticality assessment may not be the most suitable approach. Future iterations might benefit from a different architecture that are more suited for binary decisions, such as \textit{CNN-LSMTs}.

Despite its limited accuracy, the impact of this component on overall system performance is mitigated by the pipeline’s failure-based triggering condition. Since the agentic pipeline is only invoked upon baseline failure, a separate criticality classifier becomes functionally redundant—serving as a soft filter rather than a required activation.

\subsection{Evaluation of generation efficiency}\label{sec4b}
To evaluate the generation efficiency of the agentic pipeline, two metrics were measured: the Generation Time (GT) and the Token Consumption (TC). These two metrics characterize the latency and computational cost of executing the pipeline.

After running the entire pipeline on $1000$ different scenes, the average generation time was $21.36 seconds$. The mean token consumption per scene was $45,111 tokens$. The breakdown of the generation time and token consumption is shown in Table \ref{tab:gen_eff}.

\begin{table}[H]
	\centering
    	\caption{Generation Time (GT, in seconds) and Token Consumption (TC) for each agent in the pipeline, averaged over 1000 scenes.}
	\begin{tabular}{|c|c|c|c|c|c|}
		\hline
		\textbf{Metric} & \textbf{Total} & \textbf{Descriptor} & \textbf{Generator} & \textbf{Simulator}
		\\ \hline
		GT              & 21.36          & 10.24               & 3.93               & 7.19
		\\ \hline
		TC              & 45111          & 38488               & 2278               & 4345
		\\ \hline
	\end{tabular}
	\label{tab:gen_eff}
\end{table}

The average GT of the pipeline presents a challenge for deployment. This latency may be partially attributed to the use of remote API calls for LLM inference. While this makes the current setup more suitable as a proof of concept, future iterations could significantly reduce latency by using locally deployed LLMs. However, locally deployed models may introduce a trade-off in terms of generation quality, particularly if smaller or less capable models are used. Despite being relatively high, the current generation time is still acceptable in practice, as it enables the vehicle to continue autonomous operation without needing to hand over control to the driver—making it a better alternative to manual intervention in safety-critical scenarios.

Furthermore, both the GT and TC are dominated by the Descriptor, which is expected as it is the most complex agent in the pipeline. This indicates that optimizing the Descriptor’s architecture could yield the most substantial efficiency gains, though this could potentially reduce the system’s ability to generalize to novel or ambiguous scenarios.

\subsection{Evaluation of the agentic pipeline}\label{sec4c}
To evaluate the pipeline from a holistic perspective, scenarios were developed to test the entire pipeline qualitatively. The scenarios were designed using the ROS-based autonomous driving stack explained in Section \ref{sec3}. 

\begin{figure}[H]
    \centering
    \includegraphics[width = 1\linewidth]{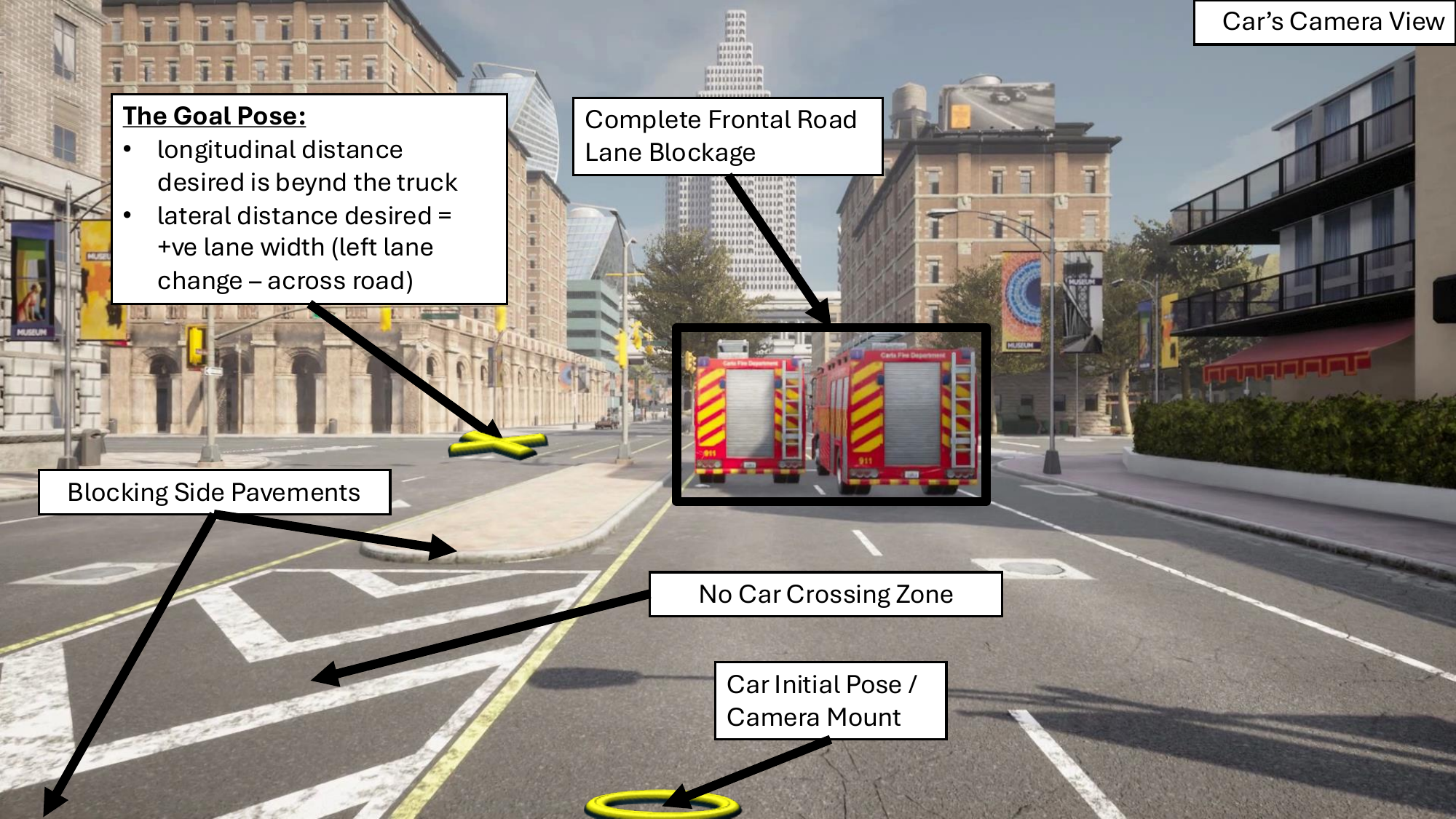}
    \caption{Vehicle's Initial Position and Initial Scene Description with Obstructions and Goal Pose}
    \label{fig:SceneDecription}
\end{figure}

In the selected scenario, the ego vehicle drives along a two-lane road and encounters a fire truck blocking its lane. The car's goal is to proceed past the obstruction and reach a navigation goal beyond the blocked segment along the opposite direction road segment (moving to the left). Fig. \ref{fig:SceneDecription} illustrates the described scenario. This is an example of lane obstruction with lane change, which is a common driving scenario. 
The car, using the baseline behavior tree provided by \texttt{Nav2}, would approach the fire truck and stop indefinitely. After a few seconds, the baseline behavior tree would return a state of failure, as the car would be unable to reach its goal, which, in the implemented architecture, triggers the agentic pipeline. 

The pipeline was tested using a simulated image of the scenario, captured from the ego vehicle’s front-facing camera. The descriptor agent correctly identified the criticality of the scene and generated a description of the situation. This was then passed to the planner agent, which generated a plan to perform a turning maneuver. 

\begin{figure*}[t]
\centering
\begin{subfigure}[t]{0.4\linewidth}
\centering
\includegraphics[width=1\linewidth]{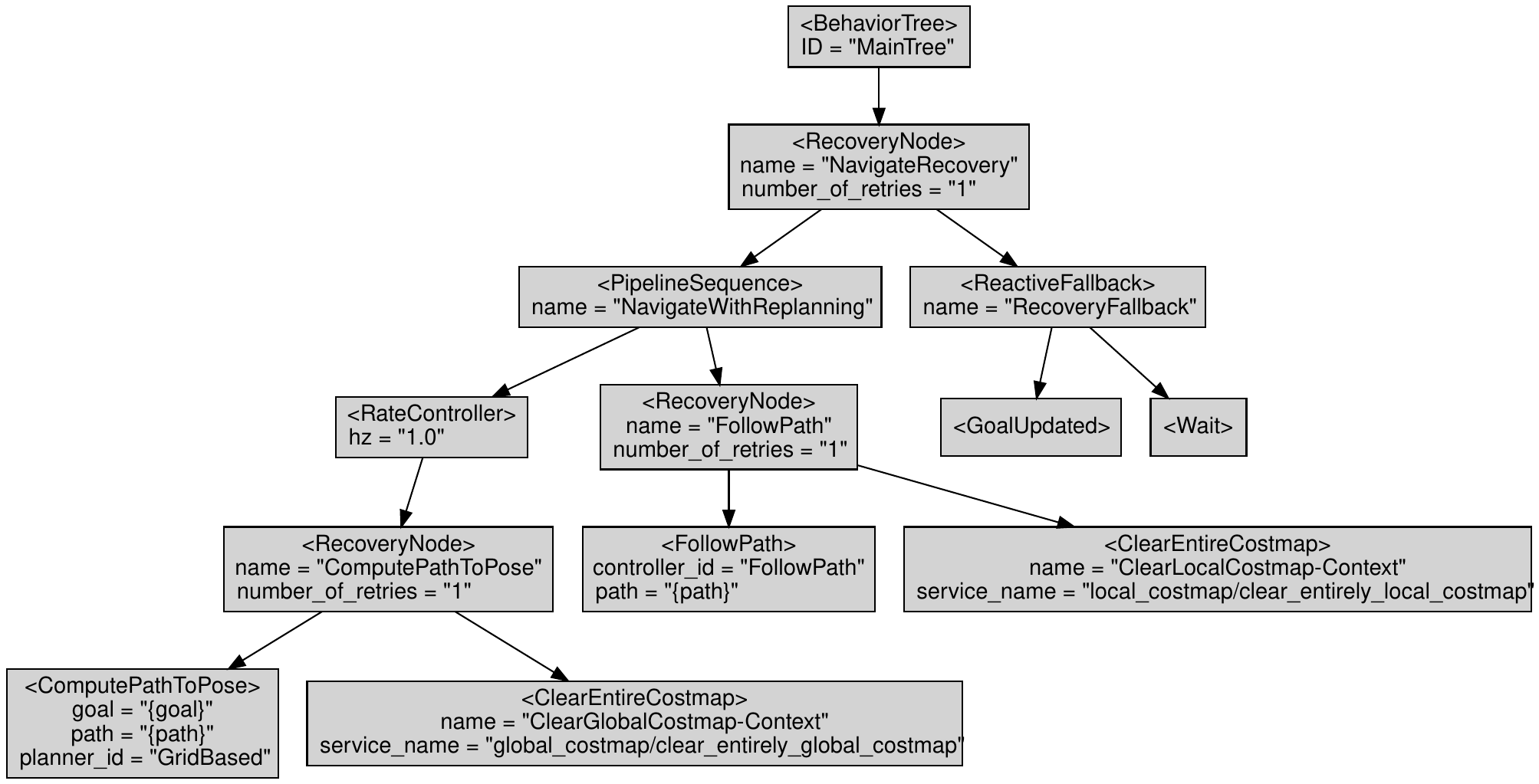}
\caption{}
\label{fig:BT_baseline}
\end{subfigure}
\begin{subfigure}[t]{0.5\linewidth}
\centering
\includegraphics[width=1\linewidth]{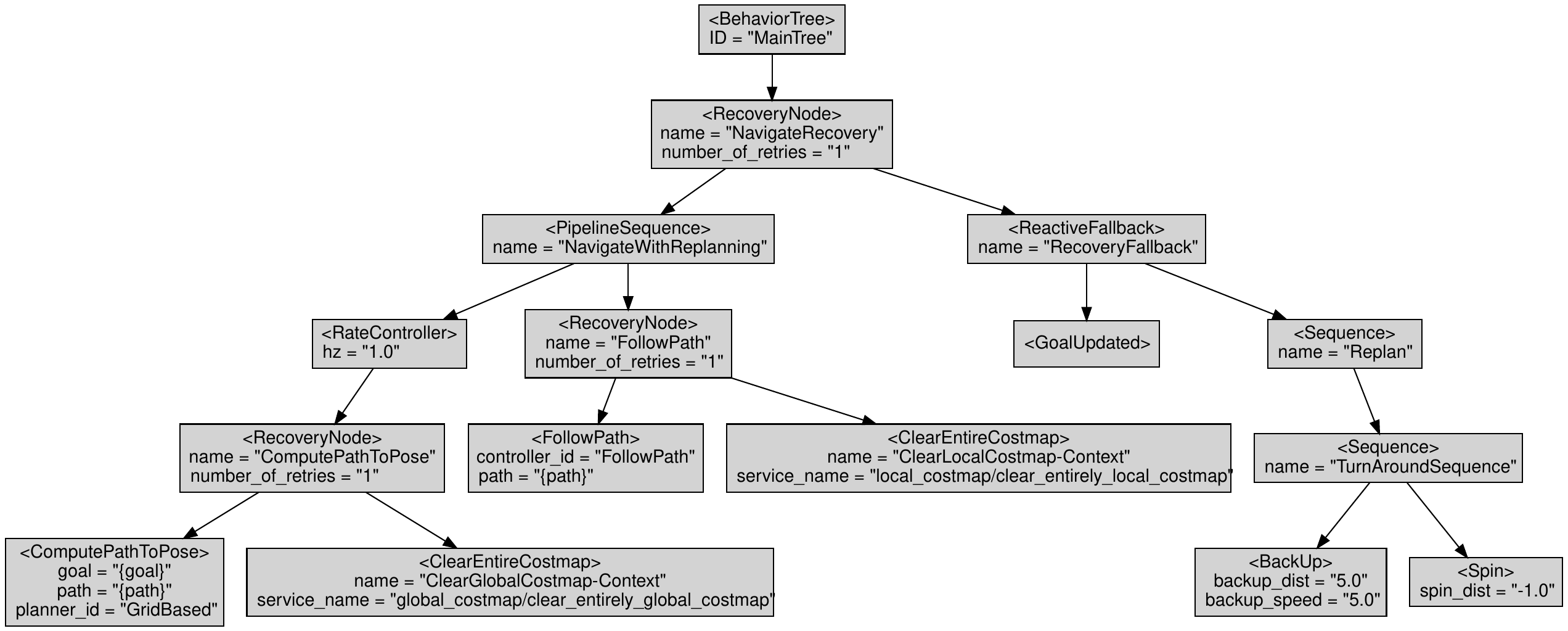}
\caption{}
\label{fig:BT_LLM}
\end{subfigure}
\caption{The Behavior Trees (a) Baseline, (b) \textit{Prompt to Pavement} Framework Generated}
\label{fig:BT}
\end{figure*}

Finally, The Generator synthesized a behavior tree corresponding to the planner’s output, which was able to successfully navigate around the fire truck and reach the navigation goal. Fig. \ref{fig:BT_baseline} illustrates the baseline BT introduced by \texttt{Nav2}, while Fig. \ref{fig:BT_LLM} illustrates the output BT from the agentic pipeline and it is clear that the proposed architecture succeeded in expending the baseline BT, adding new action nodes to resolve this obstructed scenario.

The generated behavior tree was not only valid and executable, but it also exhibited a high degree of interpretability and alignment with the planner’s output. While the behavior tree was structurally correct, some parameter tuning was required for successful execution — notably, adjusting the turn angle. This result demonstrates the pipeline’s potential for adaptive behavior planning. At the same time, it exposes key limitations in generalization and reasoning depth, which are addressed in Section~\ref{sec5}. Nevertheless, this experiment confirms the viability of an agentic pipeline that can adapt the behavior planning of a vehicle.

The final demonstrated results can be shown from the multimedia accessible from the link: \url{https://youtu.be/p0wKg-6DsoY}. This video shows the results of the same scenario using the baseline BT and how it failed and got stuck without reaching the goal. On the other hand, the LLM generated BT from the proposed framework successfully got the vehicle out of obstruction and managed to reach the goal.

\section{Conclusion and Future Recommendations}\label{sec5}
While the Prompt to Pavement framework demonstrates promising adaptability in behavior tree generation, there remain areas for improvement that warrant further exploration. The agentic pipeline is capable of generating BTs for a variety of scenes and tasks, but it struggles with complex patterns. 

The distinction between simple and complex patterns in the current pipeline is based on the structural similarity between encountered scenes. Simple patterns are defined as cases where two different scenes share the same underlying problem structure, such as encountering a static obstacle and requiring a navigation adjustment, regardless of surface-level differences like the type of obstacle or environmental context. Complex patterns, conversely, involve fundamentally different problem structures, such as navigating around a static obstacle versus navigating around a dynamic obstacle. This limitation is hypothesized to arise from the current simplicity of the agentic architecture, which may lack the capacity for deeper contextual reasoning required in novel scenarios. Provided a more intricate agentic architecture, the pipeline may be able to detect more complex patterns and generate more effective BTs.

Furthermore, the current evaluation framework relies heavily on a simulated environment, in this case \textit{CARLA}. While this allows for rapid testing and iteration, it may not fully capture the complexities and unpredictability of real-world scenarios. Future work includes developing a more grounded evaluation framework capable of assessing the performance of the generated BTs under real-world conditions.

Another area for future improvement is that the generated BTs are not formally verified, and while the current pipeline is designed to be robust against failures, there is no guarantee that it will always function correctly. The pipeline, when faced with a failure in the generated BT will prompt the human operator to intervene.
\bibliographystyle{IEEEtran}
\bibliography{sections/ref} 
\end{document}